\documentclass[sigconf, nonacm]{acmart}
\usepackage[table]{xcolor}
\usepackage{multirow}
\AtBeginDocument{%
  }

\begin{document}

\title{TIGFlow-GRPO: Trajectory Forecasting via Interaction-Aware Flow Matching and Reward-Guided Optimization}

\author{Xuepeng Jing}
\email{xuepeng_j@tju.edu.cn}
\orcid{0009-0002-6365-9518}
\affiliation{%
  \institution{Tianjin University}
  \city{Tianjin}
  \country{China}
}

\author{Wenhuan Lu}
\affiliation{%
  \institution{Tianjin University}
  \city{Tianjin}
  \country{China}
}
\author{Hao Meng}
\affiliation{%
  \institution{Tianjin University}
  \city{Tianjin}
  \country{China}
}
\author{Zhizhi Yu}
\authornote{Corresponding author.}
\affiliation{%
  \institution{Tianjin University}
  \city{Tianjin}
  \country{China}
}
\author{Jianguo Wei}
\affiliation{%
  \institution{Tianjin University}
  \city{Tianjin}
  \country{China}
}


\begin{abstract}
Human trajectory forecasting is important for intelligent multimedia systems
operating in visually complex environments, such as autonomous driving and crowd
surveillance. Although Conditional Flow Matching (CFM) has shown strong ability
in modeling trajectory distributions from spatio-temporal observations, existing
approaches still focus primarily on supervised fitting, which may leave social
norms and scene constraints insufficiently reflected in generated trajectories.
To address this issue, we propose TIGFlow-GRPO, a two-stage generative
approach that aligns flow-based trajectory generation with behavioral rules. In 
the first stage, we build a CFM-based predictor with a Trajectory-Interaction-Graph 
(TIG) module to model fine-grained visual-spatial interactions and 
strengthen context encoding. This stage captures both agent-agent and agent-scene 
relations more effectively, providing more informative conditional features for 
subsequent alignment. In the second stage, we perform Flow-GRPO post-training, 
where deterministic flow rollout is reformulated as stochastic ODE-to-SDE sampling 
to enable trajectory exploration, and a composite reward combines view-aware social 
compliance with map-aware physical feasibility. By evaluating trajectories explored 
through SDE rollout, GRPO progressively steers multimodal predictions toward behaviorally 
plausible futures. Experiments on the ETH/UCY and SDD datasets show that TIGFlow-GRPO
improves forecasting accuracy and long-horizon stability while generating
trajectories that are more socially compliant and physically feasible. These
results suggest that the proposed approach provides an effective way to connect
flow-based trajectory modeling with behavior-aware alignment in dynamic
multimedia environments.
\end{abstract}

\begin{teaserfigure}
  \includegraphics[width=\textwidth]{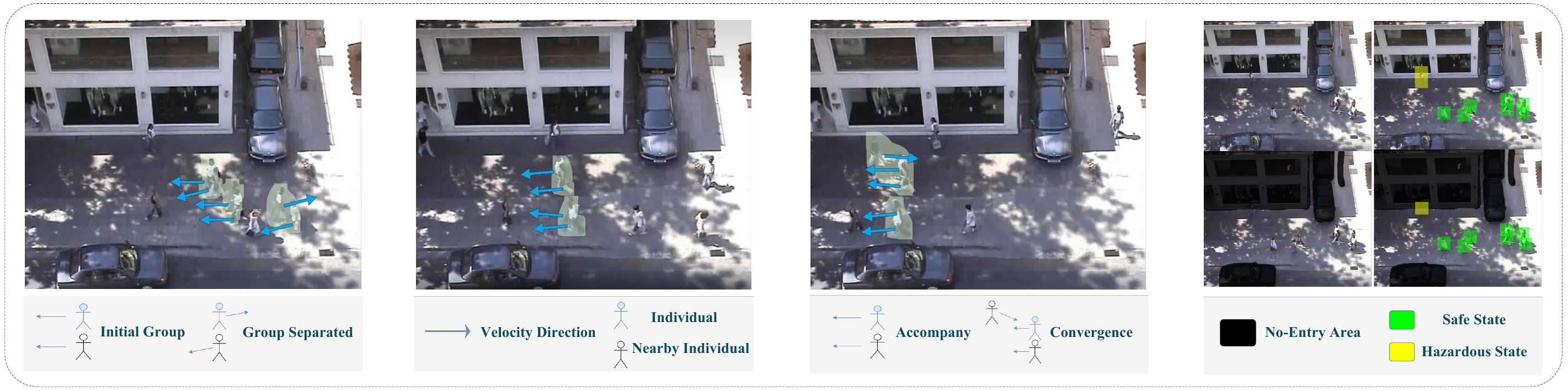}
  \caption{Representative examples of social interaction patterns and scene semantics. From left to right, 
  the teaser shows group separation, individual motion direction, accompany and convergence behaviors, and 
  semantic scene regions with different safety states, including no-entry, hazardous, and safe areas.}
  \Description{Top-view pedestrian scene examples illustrating representative interaction patterns, 
  including group separation, individual motion direction, accompany and convergence behaviors, together 
  with semantic regions labeled as no-entry, hazardous, and safe.}
  \label{fig:teaser}
\end{teaserfigure}

\maketitle

\section{Introduction}
Human trajectory forecasting is a fundamental problem in multimedia perception 
and spatio-temporal data mining~\cite{mangalam2021goals,hetzel2024reliable}. 
It supports applications such as autonomous driving, intelligent video surveillance, 
and human-robot interaction by providing predictive motion context. 
In dense and visually complex scenes, the task is to predict future pedestrian trajectories 
from observed motion histories while ensuring that the predicted paths are 
socially acceptable and physically feasible~\cite{gupta2018social,salzmann2020trajectron++}. 
This problem differs from conventional time-series forecasting because human motion is shaped by 
uncertain intentions, limited visual perception, nearby pedestrians, and 
the geometry of the surrounding scene. 
Consequently, future motion is inherently multimodal~\cite{xu2022socialvae,bae2023eigentrajectory,gu2022stochastic}: 
given the same observation history, multiple futures may be statistically plausible, 
although only a subset of them are compatible with real-world social norms and physical constraints.

A central challenge in trajectory forecasting is the discrepancy between statistical 
distribution fitting and behavior-aware generation. 
Most recent methods are trained with supervised objectives, which encourage models to 
match empirical trajectory distributions~\cite{xu2022socialvae,gu2022stochastic}. 
Although this strategy captures common motion patterns effectively, it does not directly 
optimize compliance with social conventions or rules of the scene. 
Consequently, even strong trajectory encoders and interaction modules may still produce 
implausible behaviors when the training objective fail to explicitly penalize unsafe interactions 
or invalid spatial decisions~\cite{bae2024singulartrajectory,fu2025moflow}. 
Small deviations introduced early in sequential prediction can accumulate over time and 
drive the generated sequence toward states that lack data support. 
The resulting trajectories may remain statistically plausible while still violating basic 
constraints, such as crossing obstacles or conflicting with nearby 
pedestrians~\cite{hetzel2024reliable,bahari2025certified}. 
This mismatch suggests that reliable forecasting in interactive environments requires an 
explicit mechanism for behavioral alignment~\cite{dong2025leveraging}.

Recent progress in continuous-time generative modeling provides a useful foundation for this problem. 
Models such as diffusion models and CFM demonstrate strong performance in 
high-quality generation~\cite{ho2020denoising,dhariwal2021diffusion,lipman2022flow,esser2024scaling}. 
In particular, CFM learns continuous normalizing flows through vector field regression, 
which yields stable training without simulation and supports efficient inference with 
few steps~\cite{lipman2022flow,liu2022flow}. 
These properties make CFM an attractive backbone for multimodal trajectory forecasting. 
However, this advantage should be understood as a modeling foundation rather than a complete solution, 
because the objective of CFM remains data-driven and does not inherently resolve 
behavioral alignment~\cite{fu2025moflow}. 
In parallel, RL has emerged as an effective tool for aligning generative models with complex, 
non-differentiable preferences~\cite{stiennon2020learning,ouyang2022training,fan2023reinforcement}. 
Among recent methods, GRPO is especially attractive because it estimates relative advantages 
within a group of generated samples and therefore does not require a separate 
value network~\cite{shao2024deepseekmath}. 
This design reduces memory overhead and simplifies policy updates, making GRPO well suited 
for generative modeling settings.

These observations motivate a straightforward strategy: utilizing CFM as the generative backbone 
and employing GRPO-style post-training to align the outputs with trajectory-specific behavioral constraints. 
A practical challenge is that standard CFM relies on deterministic rollout through an ODE, 
which limits the exploration of samples required by RL~\cite{lipman2022flow,liu2022flow}. 
Following recent methods of stochastic flow post-training~\cite{song2020score,albergo2025stochastic,domingo2024adjoint,liu2025flow}, 
we adopt a stochastic rollout during post-training so that reward-guided optimization can be 
conducted over generated trajectories. 
In this work, we integrate this stochastic rollout to effectively adapt Flow-GRPO for trajectory forecasting. 
This adaptation allows us to focus on instantiating the framework with interaction-aware conditioning 
and a task-specific reward that reflects social behavior and scene feasibility.

To this end, we propose TIGFlow-GRPO, a forecasting approach that integrates perception-aware 
interaction modeling with reward-guided flow post-training. 
The model includes a TIG-GAT module that captures social context in a target-centric manner. 
Rather than aggregating all nearby pedestrians indiscriminately, it selects neighbors according to 
the visual field of the target agent and refines the resulting interaction features for conditional 
generation, following the broader line of graph-based interaction modeling in trajectory 
forecasting~\cite{salzmann2020trajectron++,xu2022groupnet,xu2023eqmotion,gao2025socialmp}. 
During post-training, we optimize the stochastic flow policy using a composite reward tailored to 
trajectory prediction, which evaluates each generated trajectory under view-aware social rules and 
map constraints based on the signed distance field. 
This design encourages the generation of trajectories that remain compatible with both 
interpersonal behavior and scene geometry.

Our main contributions are summarized as follows.
\begin{itemize}
\item We introduce TIG-GAT, an interaction-aware perception module that
improves social context modeling in crowded scenes through target-centric,
view-aware neighbor reasoning and gated feature refinement. By combining
perception-aware neighbor selection with explicit local interaction modeling,
TIG-GAT provides richer conditional representations for socially complex
trajectory forecasting.
  
\item We design a composite reward for trajectory forecasting that integrates
social and physical constraints, including view-aware interaction rules and
scene-aware feasibility cues, to guide behavior-aware alignment during
post-training. This reward encourages flow-based trajectory generation to
produce futures that are more behaviorally plausible, socially consistent, and
physically feasible.
  
\item We adapt Flow-GRPO-style post-training to flow-based trajectory
forecasting by using stochastic rollout as the optimization mechanism for
GRPO. This design enables reward-guided alignment of a pretrained flow
predictor under non-differentiable behavioral constraints while preserving
compatibility with the learned flow prior.
\end{itemize}

\begin{figure*}[t]
  \centering
  \includegraphics[width=\textwidth]{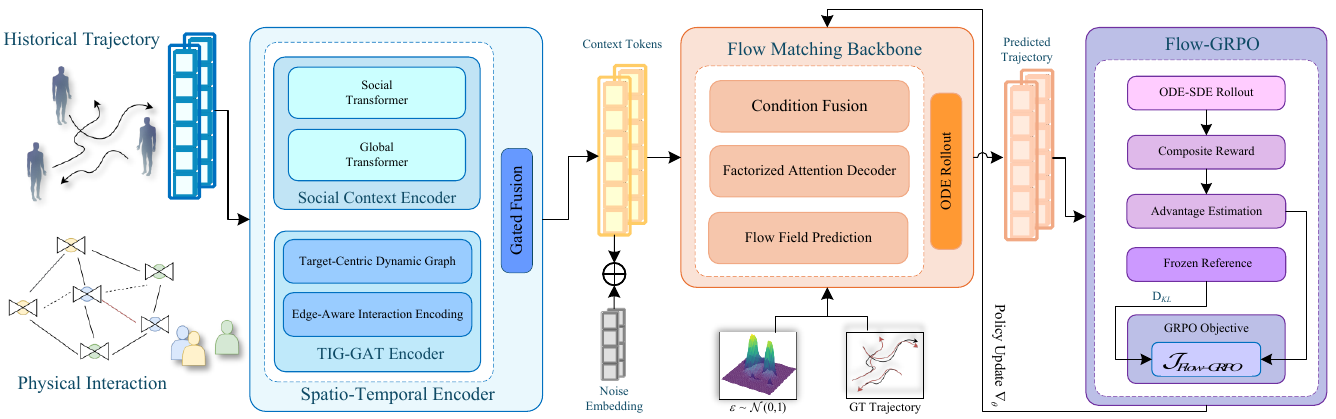}
  \caption{Overview of TIGFlow-GRPO. A spatio-temporal encoder produces context tokens from historical 
  trajectories and interaction context, the flow-matching backbone generates future trajectories through 
  ODE rollout, and Flow-GRPO post-training further aligns the predictions with social and physical constraints.}
  \Description{A left-to-right diagram of the TIGFlow-GRPO method. Historical trajectories and interaction 
  context are encoded into context tokens, which are used by a flow-matching backbone to generate future trajectories 
  through ODE rollout. A Flow-GRPO post-training module then aligns the predictions with social and physical 
  constraints using stochastic rollout, reward computation, advantage estimation, and a frozen reference branch.}
  \label{fig:framework}
\end{figure*}

\section{Related Work}
\subsection{Social Interaction Modeling in Crowds}
Early trajectory forecasting methods mainly relied on handcrafted motion priors
and shallow sequential models, which offered only limited capacity to capture
complex multi-agent interactions in crowded scenes. With the development of
graph-based reasoning, methods such as
Trajectron++~\cite{salzmann2020trajectron++},
GroupNet~\cite{xu2022groupnet}, EqMotion~\cite{xu2023eqmotion}, and
transferable GNN-based predictors~\cite{xu2022adaptive} made interaction
modeling a central component of trajectory forecasting. These methods represent
pedestrians as nodes and model their relations through spatial proximity,
relative motion, scene context, or group structure, thereby improving both
local interaction reasoning and higher-order social dependency modeling. Recent
studies further extend this direction by learning socially aware motion
patterns~\cite{gao2025socialmp} and interaction-aware stochastic
formulations~\cite{fang2025neuralized}. Even so, graph construction still
involves a trade-off: dense connectivity may introduce noisy or redundant
relations, whereas distance-based pruning may remove neighbors that are
behaviorally informative. This challenge becomes more pronounced in visually
complex scenes, where meaningful interaction cues depend not only on distance
but also on perception, visibility, and local context. To better address this
issue, TIG-GAT adopts target-centric, perception-aware neighbor selection and
temporal gating, enabling the model to retain informative interactions while
suppressing irrelevant social noise.

\subsection{Generative Trajectory Forecasting}
Recent generative forecasting methods have increasingly focused on the
intrinsic multimodality of human motion. Earlier latent-variable and
structured-representation methods, such as
SocialVAE~\cite{xu2022socialvae} and
EigenTrajectory~\cite{bae2023eigentrajectory}, improve distribution modeling
through time-dependent latent variables or low-rank motion descriptors.
Diffusion-based predictors, including MID~\cite{gu2022stochastic},
SingularTrajectory~\cite{bae2024singulartrajectory}, and
DD-MDN~\cite{hetzel2026dd}, further enhance multimodal generation by
formulating future prediction as a stochastic denoising process, while more
recent distillation-based variants aim to improve efficiency in multimodal
forecasting~\cite{jeong2025multi}. By comparison, flow-based methods learn a
continuous transformation from a simple prior to the target trajectory
distribution, offering a favorable balance between training stability and
sampling efficiency. MoFlow~\cite{fu2025moflow}, in particular, demonstrates
the potential of one-step flow matching for human trajectory forecasting.
However, existing flow-based predictors for trajectory forecasting still depend
mainly on supervised data fitting, and related flow-based generative models are
also commonly trained with supervised
objectives~\cite{polyak2024movie,gat2024discrete,esser2024scaling}. As a
result, the alignment of generated trajectories with hard physical boundaries
and complex social norms remains limited, which motivates our post-training
alignment strategy for flow-based trajectory generation.

\subsection{Policy Optimization and Constraints}
To handle non-differentiable real-world constraints, such as off-road
penalties and scene-specific map compliance, RL provides a natural
method~\cite{stiennon2020learning,ouyang2022training,shao2024deepseekmath}.
Recently, GRPO has emerged as an efficient critic-free policy optimization
method and has attracted growing attention in LLM alignment and other
generative settings
\cite{shao2024deepseekmath,fan2025online,zheng2025yingmusic}. Subsequent work
has extended GRPO-style training to continuous generative models, showing that
ODE-to-SDE conversion can introduce controllable stochasticity into otherwise
deterministic flow rollouts~\cite{xue2025advantage,sun2025f5r}. However, the use of this idea in human trajectory
forecasting is still relatively limited. Unlike image or speech generation,
motion forecasting must jointly account for multimodal diversity, positional
accuracy, map-aware feasibility, and local collision avoidance. Motivated by
this gap, TIGFlow-GRPO combines flow-based generation with GRPO and a tailored
composite reward to better align the generative backbone with visual field
rules and map constraints.

\section{Methodology}
\subsection{Problem Formulation}
Consider a visually complex crowd scene with $A$ interacting agents. For a target 
agent $i \in \{1, \ldots, A\}$, the observed trajectory over the past $T_h$ steps is 
denoted by $\mathbf{X}_i=\{\mathbf{x}_i^{-T_h+1}, \ldots, \mathbf{x}_i^0\}$, where 
$\mathbf{x}_i^t \in \mathbb{R}^2$ is the 2D position at time step $t$. We define the 
conditioning context as $\mathcal{C}_i = \{\mathbf{X}_i, \mathcal{N}_i, \mathcal{M}\}$, 
where $\mathcal{N}_i$ encodes local social interactions under visual field rules and 
$\mathcal{M}$ denotes scene topology. The goal is to model the conditional 
distribution of future motion $p(\mathbf{Y}_i \mid \mathcal{C}_i)$, where 
$\mathbf{Y}_i=\{\mathbf{y}_i^1, \ldots, \mathbf{y}_i^{T_f}\}$ is the future trajectory 
over the next $T_f$ steps. To capture multimodal uncertainty, we adopt a Conditional 
Flow Matching (CFM) method \cite{lipman2022flow,fu2025moflow}, which learns a 
time-dependent vector field $v_\theta(\mathbf{z}_t, t, \mathcal{C}_i)$ that transports 
a Gaussian prior $\mathbf{z}_0 \sim \mathcal{N}(\mathbf{0}, \mathbf{I})$ to the target 
distribution $\mathbf{z}_1$ through an Ordinary Differential Equation (ODE):
\begin{equation}
    d\mathbf{z}_t = v_\theta(\mathbf{z}_t, t, \mathcal{C}_i) dt, \quad t \in [0, 1].
    \label{eq:ode}
\end{equation}
Since standard CFM relies on supervised fitting and does not explicitly enforce 
physical or social constraints, we further adopt a two-stage approach that combines 
flow-matching generation with ODE-to-SDE reformulation and reinforcement learning 
for behavioral alignment \cite{song2020score,fan2025online,liu2025flow}.

\begin{figure*}[t]
  \centering
  \includegraphics[width=\textwidth]{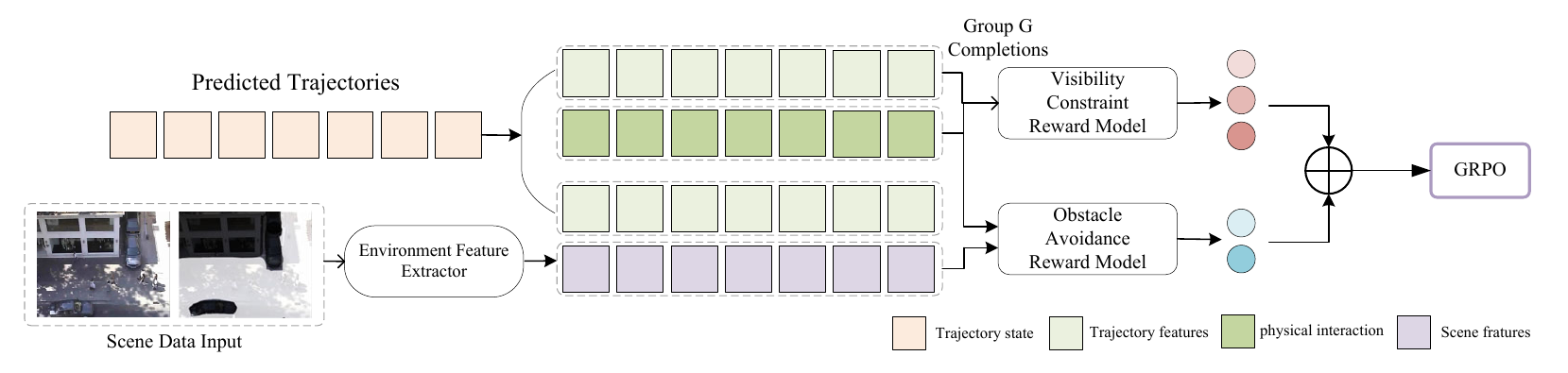}
  \caption{Flow-GRPO reward modeling and policy optimization. Predicted trajectories are organized into $G$ completions and combined with environment features extracted from scene data. A visibility-constraint reward model and an obstacle-avoidance reward model evaluate each completion, and their outputs are aggregated for GRPO optimization.}
  \Description{A pipeline diagram of the Flow-GRPO module. Predicted trajectories enter from the left, while scene data are processed by an environment feature extractor. The resulting group of completions contains trajectory states, trajectory features, physical interaction features, and scene features. These completions are evaluated by a visibility-constraint reward model and an obstacle-avoidance reward model, whose outputs are combined and passed to GRPO.}
  \label{fig:flow-grpo}
\end{figure*}

\subsection{Overall Network Architecture}
Figure~\ref{fig:framework} illustrates the two-stage TIGFlow-GRPO approach.
During pretraining, a dual-branch Spatio-Temporal Encoder extracts conditional
features from historical trajectories and interaction graphs. The implicit
Social Context Encoder captures coarse crowd dependencies through social and
global self-attention \cite{vaswani2017attention}, whereas the TIG-GAT Encoder
models target-centric local interactions via dynamic graph construction and
edge-aware interaction encoding. Gated Fusion combines the two branches into
context tokens that summarize global context and local interaction constraints.
Conditioned on these tokens and a Gaussian noise embedding
($\boldsymbol{\epsilon} \sim \mathcal{N}(\mathbf{0},\mathbf{I})$), the Flow
Matching Backbone predicts the flow field and performs ODE rollout to generate
future trajectories~\cite{sun2025f5r}. At this stage, the model is trained with the standard
flow-matching objective under future supervision.

During post-training, Flow-GRPO \cite{liu2025flow} reformulates the deterministic
rollout as an ODE-to-SDE sampling process
\cite{song2020score,albergo2025stochastic}. The sampled trajectories are
evaluated with a composite reward defined on the rollout and the scene
conditions, converted into group-relative advantages
\cite{shao2024deepseekmath}, and optimized with a GRPO objective regularized by
a frozen reference policy \cite{ouyang2022training,liu2025flow}. The
post-training objective depends only on sampled trajectories and scene context.
The resulting policy update is fed back into the stochastic rollout, improving
behavioral alignment while preserving multimodal diversity.

\subsection{TIG-GAT Spatio-Temporal Feature Module}
We use TIG-GAT to encode target-centric motion and local multi-agent
interactions, while estimating the interaction complexity around the target
agent.Unlike conventional graph models built on dense social
graphs~\cite{xu2022adaptive,gu2022stochastic,hetzel2024reliable}, TIG-GAT
adopts a target-centric dynamic graph. A cascaded Social
Transformer~\cite{vaswani2017attention,jiang2023motiondiffuser} first encodes
the augmented historical states into an implicit social context
$\mathbf{z}_i^{\mathrm{soc}}$. For explicit interaction modeling, TIG-GAT then
selects only the neighbors most relevant to the target agent. A field-of-view
criterion defines the binary visibility mask
$ m_{ij}^{t} = \mathbf{1}\big(\angle(\mathbf{v}_i^{t}, \Delta \mathbf{p}_{ij}^{t}) \leq \theta_{\mathrm{FoV}} \big) $,
and the dynamic neighbor set $\tilde{\mathcal{N}}_i$ is constructed from
accumulated visibility and spatial proximity over the historical window $W$:
\begin{equation}
    \tilde{\mathcal{N}}_i = \mathop{\mathrm{TopK}}_{j \in \mathcal{A} \setminus \{i\}} \sum_{\tau=-W+1}^{0} m_{ij}^{\tau} \cdot \phi_{\mathrm{sel}}\big(\|\Delta\mathbf{p}_{ij}^{\tau}\|_2\big).
    \label{eq:neighbor_selection}
\end{equation}

Based on $\tilde{\mathcal{N}}_i$, we build a local interaction graph and
introduce an Edge-Aware Gated Attention mechanism to alleviate the signal
dilution of Softmax-based attention. Each edge is parameterized by relative kinematic features
$\mathbf{e}_{ij}^{t} = \mathrm{MLP}_{\mathrm{edge}}\big([\Delta \mathbf{p}_{ij}^{t} \parallel \Delta \mathbf{v}_{ij}^{t} \parallel \|\Delta \mathbf{p}_{ij}^{t}\|_2 \parallel \cos\theta_{ij}^{t}]\big)$.
Message passing uses independent spatial gates:
\begin{equation}
    \begin{aligned}
        \mathbf{g}_{ij}^{t} &= \sigma\Big(\mathrm{MLP}_{\mathrm{spa}}\big([\mathbf{h}_i^{t} \parallel \mathbf{h}_j^{t} \parallel \mathbf{e}_{ij}^{t}]\big)\Big), \\
        \mathbf{m}_i^{t} &= \sum_{j \in \tilde{\mathcal{N}}_i} \mathbf{g}_{ij}^{t} \odot \big(\mathbf{W}_V\mathbf{h}_j^{t} + \mathbf{W}_E\mathbf{e}_{ij}^{t}\big),
    \end{aligned}
    \label{eq:message_passing}
\end{equation}
where $\sigma(\cdot)$ denotes the Sigmoid activation function. This mechanism
allows multiple relevant neighbors to contribute without competing through a
normalized attention distribution~\cite{xiao2025srefiner}.

After spatial aggregation, Temporal Gated Pooling summarizes the frame-wise
physical features into a compact physical representation
$\mathbf{h}_i^{\mathrm{phy}}$, while the accumulated spatial gate activations
provide a scalar interaction-strength estimate $s_i$. Finally, a
strength-modulated gated residual fuses $\mathbf{h}_i^{\mathrm{phy}}$ with the
implicit context $\mathbf{z}_i^{\mathrm{soc}}$ to produce the context token
$\mathbf{c}_i$:
\begin{equation}
    \begin{aligned}
        \mathbf{g}_i &= \sigma\Big(\mathrm{MLP}_{\mathrm{fuse}}\big([\mathbf{z}_i^{\mathrm{soc}} \parallel \mathbf{h}_i^{\mathrm{phy}} \parallel s_i]\big)\Big), \\
        \mathbf{c}_i &= \mathbf{z}_i^{\mathrm{soc}} + \lambda s_i \mathbf{g}_i \odot \tanh\big(\mathbf{W}_{\mathrm{phy}}\mathrm{LN}(\mathbf{h}_i^{\mathrm{phy}})\big).
    \end{aligned}
    \label{eq:gated_fusion}
\end{equation}

By combining implicit social context with explicit local interaction modeling,
TIG-GAT produces the context representation used in the subsequent generative
stages.

\subsection{Flow-GRPO Post-Training}
After supervised pretraining, we use Flow-GRPO as a post-training stage for
trajectory forecasting~\cite{liu2025flow}. Standard flow rollout is deterministic
under a fixed scene condition~\cite{lipman2022flow,liu2022flow} and therefore
cannot provide the exploration needed to compare multiple plausible futures.
We therefore preserve the pretrained flow prior and introduce stochasticity only
during post-training, allowing the model to sample several candidate futures
from the same observation and optimize them with trajectory-level
rewards~\cite{fan2025online}. In our implementation, we use all rollout steps
during optimization.

Let $v_{\theta}(\mathbf{y}_t, t, \mathbf{c}_i)$ denote the pretrained velocity
field for latent state $\mathbf{y}_t$ at time $t$ under scene context
$\mathbf{c}_i$. We keep the same Gaussian OT-CFM path as in pretraining,
$\mathbf{y}_t = t\mathbf{y}_1 + (1-t)\boldsymbol{\xi}$ with
$\boldsymbol{\xi} \sim \mathcal{N}(\mathbf{0}, \mathbf{I})$, so that
post-training remains consistent with the learned flow prior. Under this
Gaussian path, the score can be recovered directly from the velocity
field~\cite{albergo2025stochastic}. To avoid numerical instability near the
endpoints, we use
$\bar t = \mathrm{clip}(t, \tau_{\min}, 1-\tau_{\min})$. The recovered score
and diffusion coefficient are:
\begin{equation}
\begin{aligned}
\mathbf{s}_{\theta}(\mathbf{y}_t,\bar t,\mathbf{c}_i)
&=
\frac{\bar t\,v_{\theta}(\mathbf{y}_t,\bar t,\mathbf{c}_i)-\mathbf{y}_t}
{1-\bar t}, \\
g(\bar t)
&=
\eta \sqrt{\frac{1-\bar t}{\bar t}}.
\end{aligned}
\label{eq:score_and_diffusion}
\end{equation}

Using the recovered score, we replace deterministic ODE rollout with the
following stochastic transition~\cite{song2020score,albergo2025stochastic,liu2025flow}:
\begin{equation}
\begin{aligned}
\mu_t
&=
\mathbf{y}_t
+
\Big[
v_{\theta}(\mathbf{y}_t,\bar t,\mathbf{c}_i)
+
\tfrac{1}{2} g(\bar t)^2
\mathbf{s}_{\theta}(\mathbf{y}_t,\bar t,\mathbf{c}_i)
\Big]\Delta t, \\
\mathbf{y}_{t+\Delta t}
&=
\mu_t + \sigma_t \boldsymbol{\varepsilon}_t,
\qquad
\sigma_t = g(\bar t)\sqrt{\Delta t},
\qquad
\boldsymbol{\varepsilon}_t \sim \mathcal{N}(\mathbf{0}, \mathbf{I}).
\end{aligned}
\label{eq:sde_transition}
\end{equation}

This transition defines a Gaussian policy at each rollout step and thus enables
direct evaluation of stepwise log-probabilities~\cite{xue2025advantage}. Under
the same scene condition, we sample $G$ rollouts from a shared initial noise and
evaluate each complete trajectory with the composite reward defined above.

For the $g$-th rollout, we compute the group-relative advantage
$A_g = (\mathcal{R}_g-\mu_{\mathcal{R}})/(\sigma_{\mathcal{R}}+\epsilon_{\mathrm{adv}})$~\cite{shao2024deepseekmath,sun2025f5r}.
We further define the stepwise policy ratio
$r_{g,t}(\theta)=\exp(\log p_{\theta}-\log p_{\mathrm{old}})$ and its clipped
version
$\bar r_{g,t}(\theta)=\mathrm{clip}(r_{g,t}(\theta),1-\epsilon_{\mathrm{clip}},
1+\epsilon_{\mathrm{clip}})$. To keep the updated policy close to the
pretrained predictor, we introduce a frozen reference model and penalize the
discrepancy between Gaussian transition means, which corresponds to the
Gaussian KL term under a shared variance
schedule~\cite{ouyang2022training,liu2025flow}. The final objective is:
\begin{equation}
\begin{aligned}
\mathcal{L}_{\text{Flow-GRPO}}
=
\frac{1}{G|\mathcal{S}|}
\sum_{g=1}^{G}\sum_{t\in\mathcal{S}}
\Big[
&-\min\big(
r_{g,t}(\theta)A_g,\,
\bar r_{g,t}(\theta)A_g
\big) \\
&+
\beta
\frac{\|\mu_{\theta,g,t}-\mu_{\mathrm{ref},g,t}\|_2^2}
{2\sigma_t^2}
\Big].
\end{aligned}
\label{eq:grpo_objective}
\end{equation}

This design makes Flow-GRPO a behavior-aware post-training method for
trajectory forecasting, improving social and physical consistency while
preserving the multimodal prior learned through supervised flow
matching~\cite{liu2025flow}.

\subsection{Reward Function Design}

\subsubsection{View-Aware Social Reward}
To avoid treating all neighboring interactions equally, we define a view-aware
social reward based on direction-sensitive geometry~\cite{gao2025socialmp}. For
the $g$-th predicted trajectory of target agent $i$, we first transform the
prediction into the absolute coordinate system,
$\hat{\mathbf{y}}_{i,g}^{\,t,\mathrm{abs}} = \hat{\mathbf{y}}_{i,g}^{\,t} + \mathbf{x}_i^0$.
We then construct frame-wise heading vectors:
\begin{equation}
    \mathbf{h}_{i,g}^t =
    \begin{cases}
        \hat{\mathbf{y}}_{i,g}^{\,1,\mathrm{abs}} - \mathbf{x}_i^0, & t=1, \\
        \hat{\mathbf{y}}_{i,g}^{\,t,\mathrm{abs}} - \hat{\mathbf{y}}_{i,g}^{\,t-1,\mathrm{abs}}, & t>1.
    \end{cases}
    \label{eq:view_heading}
\end{equation}

Based on the local heading, we identify relevant neighbors. Let
$\mathbf{r}_{ij,g}^t = \hat{\mathbf{y}}_{j,g}^{\,t,\mathrm{abs}} - \hat{\mathbf{y}}_{i,g}^{\,t,\mathrm{abs}}$
denote the relative position vector~\cite{bae2024singulartrajectory}. The
viewing angle $\theta_{ij,g}^t$ is computed only for neighbors within an
interaction radius $R$ and with valid heading support
($\|\mathbf{h}_{i,g}^t\|_2 > \epsilon$):
\begin{equation}
    \theta_{ij,g}^t = \arccos\left( \frac{\langle \mathbf{h}_{i,g}^t,\mathbf{r}_{ij,g}^t\rangle}{\|\mathbf{h}_{i,g}^t\|_2\,\|\mathbf{r}_{ij,g}^t\|_2+\epsilon} \right).
    \label{eq:view_angle}
\end{equation}

We divide interactions into strong-view ($s$), weak-view ($w$), and rear-view
($r$) regions. Let
$m_{ij,g}^t = \mathbf{1}\big( \|\mathbf{r}_{ij,g}^t\|_2 \le R \land \|\mathbf{h}_{i,g}^t\|_2 > \epsilon \big)$
denote the valid-neighbor mask. The regional masks are defined by angular
thresholds $(\phi_s, \phi_w)$:
\begin{equation}
    \begin{aligned}
        \mathcal{M}_{ij,g}^{s,t} &= m_{ij,g}^t \cdot \mathbf{1}(\theta_{ij,g}^t \le \phi_s), \\
        \mathcal{M}_{ij,g}^{w,t} &= m_{ij,g}^t \cdot \mathbf{1}(\phi_s < \theta_{ij,g}^t \le \phi_w), \\
        \mathcal{M}_{ij,g}^{r,t} &= m_{ij,g}^t \cdot \mathbf{1}(\theta_{ij,g}^t > \phi_w).
    \end{aligned}
    \label{eq:view_mask}
\end{equation}

Using the predicted pairwise distance
$\hat{d}_{ij,g}^t = \|\mathbf{r}_{ij,g}^t\|_2$ and its temporal change
$\Delta d_{ij,g}^t = \hat{d}_{ij,g}^t - \hat{d}_{ij,g}^{t-1}$, with
$\hat{d}_{ij,g}^{0} = \|\mathbf{x}_j^0 - \mathbf{x}_i^0\|_2$, we assign
region-specific penalties. The strong-view region penalizes unsafe proximity,
the weak-view region balances close contact and over-avoidance~\cite{sun2025f5r,liu2025flow},
and the rear-view region only mildly penalizes excessive separation. With a
scene-density normalization factor $\gamma_i$, the resulting reward is:
\begin{equation}
    \begin{aligned}
        r_{i,g}^{\mathrm{sv}} = -\gamma_i \operatorname{Avg}_{j,t} \Big( &w_s \mathcal{M}_{ij,g}^{s,t} [\delta_s - \hat{d}_{ij,g}^t]_+^2 \\
        &+ w_{wc} \mathcal{M}_{ij,g}^{w,t} [\delta_w - \hat{d}_{ij,g}^t]_+^2 \\
        &+ w_{wo} \mathcal{M}_{ij,g}^{w,t} [\Delta d_{ij,g}^t - m_w]_+^2 \\
        &+ w_r \mathcal{M}_{ij,g}^{r,t} [\Delta d_{ij,g}^t - m_r]_+^2 \Big).
    \end{aligned}
    \label{eq:view_reward}
\end{equation}
This reward introduces an explicit direction-sensitive social constraint within
Flow-GRPO.

\begin{table*}[t!]
    \centering
    \caption{Quantitative comparison of best-of-$K$ ADE$_{\min}$ / FDE$_{\min}$ on the ETH/UCY benchmark, where the minimum error is reported over $K$ predicted trajectories. Bold and underlined entries denote the best and second-best results, respectively. Lower values indicate better forecasting accuracy, and TIGFlow-GRPO achieves the best overall performance.}
    \label{tab:quantitative_ethucy}
    \resizebox{\textwidth}{!}{
    \begin{tabular}{c|cccccccc|c}
    \toprule
    \raisebox{1.6ex}{Datasets} 
    & \shortstack[c]{GroupNet\\\cite{xu2022groupnet}}
    & \shortstack[c]{MID\\\cite{gu2022stochastic}}
    & \shortstack[c]{EqMotion\\\cite{xu2023eqmotion}}
    & \shortstack[c]{EigenTraj\\\cite{bae2023eigentrajectory}}
    & \shortstack[c]{LED\\\cite{mao2023leapfrog}}
    & \shortstack[c]{SingularTraj\\\cite{bae2024singulartrajectory}}
    & \shortstack[c]{MoFlow\\\cite{fu2025moflow}}
    & \shortstack[c]{DD-MDN\\\cite{hetzel2026dd}}
    & \shortstack[c]{\textbf{TIGFlow-GRPO}\\{\scriptsize Ours}} \\ 
    \midrule
    ETH     & 0.46/0.73 & 0.39/0.66 & 0.40/0.62 & \underline{0.37}/\underline{0.54} & 0.39/0.58 & \textbf{0.35}/\textbf{0.43} & 0.40/0.60 & \textbf{0.35}/\underline{0.54} & 0.39/\underline{0.55} \\
    HOTEL   & 0.15/0.25 & 0.13/0.22 & \textbf{0.11}/0.18 & 0.13/0.19 & \textbf{0.11}/\underline{0.17} & 0.13/0.20 & \underline{0.12}/0.18 & 0.13/0.18 & \textbf{0.11}/\textbf{0.16} \\
    UNIV    & 0.26/0.49 & \textbf{0.22}/0.45 & \underline{0.23}/0.42 & 0.24/0.40 & 0.26/0.44 & 0.26/0.45 & \underline{0.23}/0.40 & 0.24/0.42 & \textbf{0.22}/\textbf{0.38} \\
    ZARA1   & 0.21/0.39 & \underline{0.17}/0.30 & 0.18/0.34 & 0.19/0.33 & 0.18/\textbf{0.26} & 0.19/0.33 & \underline{0.17}/0.29 & \underline{0.17}/\underline{0.28} & \textbf{0.16}/\textbf{0.26} \\
    ZARA2   & 0.17/0.33 & \underline{0.13}/0.27 & \underline{0.13}/0.25 & 0.14/0.24 & \underline{0.13}/\underline{0.22} & 0.15/0.26 & \underline{0.13}/\underline{0.22} & \textbf{0.12}/\textbf{0.21} & \textbf{0.12}/\textbf{0.21} \\ 
    \midrule
    \rowcolor{gray!20} AVG     & 0.25/0.44 & \underline{0.21}/0.38 & 0.22/0.36 & \underline{0.21}/0.34 & \underline{0.21}/0.33 & 0.22/0.33 & \underline{0.21}/0.34 & \textbf{0.20}/\underline{0.32} & \textbf{0.20}/\textbf{0.31} \\ 
    \bottomrule
    \end{tabular}}
\end{table*}

\subsubsection{Map-Aware Semantic Reward and Fusion}
For scenes with semantic maps, behaviorally plausible trajectories should also
respect the static structure of the environment
\cite{salzmann2020trajectron++,gao2025socialmp}. We therefore define a
map-aware semantic reward that incorporates walkable-area and obstacle priors
into the reinforcement learning pipeline, complementing dynamic interaction
constraints without modifying the flow backbone.

For the $g$-th sampled future, we project the predicted trajectory in the
absolute coordinate system onto the 2D map plane through the inverse rotation
matrix $R_i^{-1}$ of the focal agent and the scene homography $H_i$. To obtain
continuous boundary feedback rather than discrete occupancy
labels~\cite{dong2025leveraging,bae2024singulartrajectory}, we precompute a
Signed Distance Field (SDF), denoted by $S_i(\cdot)$, which returns the
shortest clearance from a spatial point to the nearest obstacle boundary. The
homogeneous projection and the resulting map-clearance evaluation are:
\begin{equation}
    \hat{d}_{i,g}^{\,t} = S_i\left( \Pi\big( H_i^{-1} [ (R_i^{-1}\hat{\mathbf{y}}_{i,g}^{\,t,\mathrm{abs}})^\top, 1 ]^\top \big) \right),
    \label{eq:map_projection}
\end{equation}
where $\Pi(\cdot)$ denotes homogeneous coordinate normalization. For the
observed history $\{\mathbf{x}_i^{\,k,\mathrm{abs}}\}_{k \in \mathcal{O}_i}$, we
compute the corresponding clearances $d_{i}^{\mathrm{obs},k}$ through the same
projection.

This formulation yields a continuous penalty for obstacle penetration and
unsafe proximity to scene boundaries. To avoid over-penalizing trajectories
that pass through narrow corridors near boundaries, we adopt a
relative-clearance formulation. We define the history-conditioned baseline map
risk as
$\hat{q}_{i}^{(\mathrm{obs})} = \operatorname{Avg}_{k \in \mathcal{O}_i} [\delta_{\mathrm{map}} - d_{i}^{\mathrm{obs},k}]_+^2$.
The map-aware reward penalizes only the excess risk introduced by the
prediction:
\begin{equation}
    r_{i,g}^{\mathrm{map}} = - \Big[ \operatorname{Avg}_{t} [\delta_{\mathrm{map}} - \hat{d}_{i,g}^{\,t}]_+^2 - \hat{q}_{i}^{(\mathrm{obs})} \Big]_+.
    \label{eq:map_reward}
\end{equation}
This reward encourages predicted trajectories to remain physically feasible
with respect to the local scene geometry implied by the observed motion
context~\cite{liu2025egotraj}.

Finally, we combine the map-aware reward, the view-aware social reward, and
the regularizers for trajectory accuracy ($r_{i,g}^{\mathrm{acc}}$) and motion
smoothness ($r_{i,g}^{\mathrm{sm}}$) into the unified reward for Flow-GRPO
alignment:
\begin{equation}
    \mathcal{R}_{i,g} = \omega_{\mathrm{sv}} r_{i,g}^{\mathrm{sv}} + \omega_{\mathrm{map}} r_{i,g}^{\mathrm{map}} + \omega_{\mathrm{acc}} r_{i,g}^{\mathrm{acc}} + \omega_{\mathrm{sm}} r_{i,g}^{\mathrm{sm}}.
    \label{eq:final_reward}
\end{equation}
Dataset-specific reward weights, post-training schedules, and runtime
statistics are provided in the implementation details.

\section{Experiments}
\subsection{Datasets and Evaluation Settings}
We evaluate the proposed method on two widely used public benchmarks: the
ETH/UCY pedestrian
dataset~\cite{lerner2007crowds,pellegrini2009you} and the Stanford Drone
Dataset (SDD)~\cite{robicquet2016learning}. In all experiments, trajectories
are resampled to 2.5 Hz. We use 8 observed frames (3.2 seconds) to predict the
next 12 future frames (4.8 seconds).

\subsubsection{ETH/UCY Benchmark}
ETH/UCY consists of five pedestrian scenes, namely ETH, HOTEL, UNIV, ZARA1,
and ZARA2, and is widely used to evaluate socially aware trajectory forecasting
in crowded environments. Following the standard protocol, we adopt leave-one-
out cross-validation, training on four scenes and testing on the remaining one.
This benchmark is well suited to evaluating the contributions of TIG-GAT and
the view-aware social reward.

\subsubsection{Stanford Drone Dataset (SDD)}
Captured by drones over a university campus, SDD contains dense crowds in
structurally complex environments. Its strong physical boundary constraints
make it a suitable benchmark for evaluating the map-aware semantic reward. We
precompute continuous Signed Distance Fields (SDFs) from the raw semantic scene
images and report all prediction errors in pixels.

\subsection{Experimental Setup}
\subsubsection{Baselines}
We compare TIGFlow-GRPO with a representative set of trajectory forecasting
baselines spanning multiple modeling paradigms. These baselines include
graph-based interaction models, such as
GroupNet~\cite{xu2022groupnet} and EqMotion~\cite{xu2023eqmotion}, as well as
generative predictors based on diffusion or flow matching, including
MID~\cite{gu2022stochastic},
SingularTrajectory~\cite{bae2024singulartrajectory},
MoFlow~\cite{fu2025moflow}, and the uncertainty-aware
DD-MDN~\cite{hetzel2026dd}. Since some baselines operate under different input
settings, especially with respect to the availability of semantic map cues, we
treat MoFlow~\cite{fu2025moflow} as the primary architecture-aligned baseline.
This comparison highlights the contributions of TIG-GAT and Flow-GRPO beyond
standard supervised flow-matching training, while the broader baseline set
places our method in the context of existing trajectory forecasting paradigms.

\subsubsection{Evaluation Metrics}
We evaluate multimodal coverage and overall distribution quality using two
groups of displacement metrics over $K=20$ generated samples: minimum errors
($\mathrm{ADE}_{\min}$, $\mathrm{FDE}_{\min}$) and average errors
($\mathrm{ADE}_{\mathrm{avg}}$, $\mathrm{FDE}_{\mathrm{avg}}$). The minimum
metrics compute the $L_2$ distance to the closest prediction, whereas the
average metrics measure the mean error across all generated samples. Because
the average metrics are more sensitive to low-quality or implausible samples,
they provide a more informative view of overall distribution quality. In
addition, we report a collision rate, denoted by $\mathrm{Col}$, to directly
evaluate social compliance. A prediction is counted as colliding if the
pairwise distance between any two agents falls below a predefined safety
threshold at any future step within the evaluation horizon. Lower
$\mathrm{Col}$ indicates better collision avoidance and complements the
displacement-based metrics.

\subsubsection{Implementation Details}
Our method is implemented in PyTorch and trained on a single NVIDIA RTX 3090
GPU. During supervised pretraining, TIG-GAT adopts a field-of-view angle of
$\theta_{\mathrm{FoV}} = 120^\circ$ for interaction-aware neighbor selection,
and the CFM backbone is optimized with AdamW using an initial learning rate of
$1 \times 10^{-4}$. During Flow-GRPO post-training, deterministic ODE rollout
is replaced by the proposed stochastic SDE transition, and $G=4$ trajectories
are sampled for each condition to compute group-relative advantages and optimize
the model with the composite reward. All post-training experiments are
conducted under the same single-GPU setting. For reproducibility, we provide
the dataset-specific preprocessing protocol, including frame rate, resampling
procedure, and coordinate convention for each benchmark, together with the
reward definitions, corresponding weights, and the post-training schedule.

\subsection{Comparisons with State-of-the-Art Methods}

\begin{table}
    \centering
    \caption{Quantitative comparison of best-of-$K$ ADE$_{\min}$ and FDE$_{\min}$ on the Stanford Drone Dataset (SDD), reported in pixels. Bold and underlined entries denote the best and second-best results, respectively.}
    \label{tab:quantitative_sdd}
    \vspace{-0.5em}
    \renewcommand{\arraystretch}{1.1} 
    \setlength{\tabcolsep}{5pt} 
    \begin{tabular}{c|cc}
    \midrule
    Method & ADE & FDE \\ 
    \midrule
    SocialGAN~\cite{gupta2018social}    & 27.23 & 41.44 \\
    Y-net~\cite{mangalam2021goals}        & 11.51 & 20.24 \\
    GroupNet~\cite{xu2022groupnet}     & 9.31  & 16.11 \\
    MID~\cite{gu2022stochastic}          & 9.73  & 15.32 \\
    EigenTraj~\cite{bae2023eigentrajectory}    & 8.05  & 13.25 \\
    LED~\cite{mao2023leapfrog}          & 8.49  & 12.67 \\
    MoFlow~\cite{fu2025moflow}       & \underline{7.63}  & \underline{12.25} \\ 
    \midrule
    \rowcolor{gray!20} \textbf{TIGFlow-GRPO} & \textbf{7.37} & \textbf{11.67} \\ 
    \bottomrule
    \end{tabular}
    \vspace{-0.5em}
\end{table}

\subsubsection{Performance on ETH/UCY Benchmark}
Table~\ref{tab:quantitative_ethucy} reports the quantitative results on
ETH/UCY. TIGFlow-GRPO achieves the best overall average $\mathrm{ADE}$ and
$\mathrm{FDE}$ among the compared methods, reaching 0.20 and 0.31,
respectively. Relative to MoFlow, the gains are consistent across all subsets
and are more visible in interaction-intensive scenes such as ZARA1 and UNIV.
These results indicate that TIG-GAT and Flow-GRPO improve prediction quality in
socially complex scenes.

\subsubsection{Performance on SDD Benchmark}
To assess performance in scenes with strong physical constraints, we report the
results on SDD in Table~\ref{tab:quantitative_sdd}. Under the standard
image-plane evaluation setting, TIGFlow-GRPO reduces the pixel-space
$\mathrm{ADE}$ to 7.37 and the $\mathrm{FDE}$ to 11.67. Unlike ETH/UCY, SDD is
strongly shaped by complex topological boundaries and non-walkable areas. The
improvement over MoFlow, particularly the reduction of $\mathrm{FDE}$ from 12.25
to 11.67, is consistent with the role of the map-aware semantic reward in
reducing boundary-crossing predictions.

\begin{table}
    \centering
    \caption{Performance degradation over extended prediction horizons exemplified on ETH, measured by ADE$_{\text{avg}}$ and FDE$_{\text{avg}}$. TIGFlow-GRPO consistently outperforms the MoFlow baseline across all time steps.}    \label{tab:eth_avg}
    \renewcommand{\arraystretch}{1.2} 
    \setlength{\tabcolsep}{10pt} 
    \setlength{\aboverulesep}{0pt}
    \setlength{\belowrulesep}{0pt}
    \vspace{-0.5em}
    \resizebox{\linewidth}{!}{
    \begin{tabular}{c|cc|cc}
    \hline
    ETH & \multicolumn{2}{c|}{MoFlow~\cite{fu2025moflow}} & \multicolumn{2}{c}{\textbf{TIGFlow-GRPO}} \\ 
    \cmidrule(lr){1-1} \cmidrule(lr){2-3} \cmidrule(lr){4-5} 
    Time & ADE$_{\text{avg}}$ & FDE$_{\text{avg}}$ & ADE$_{\text{avg}}$ & FDE$_{\text{avg}}$ \\ 
    \hline
    1.2s & 0.33 & 0.52 & \textbf{0.29} & \textbf{0.46} \\
    2.4s & 0.66 & 1.25 & \textbf{0.59} & \textbf{1.11} \\
    3.6s & 1.05 & 2.15 & \textbf{0.94} & \textbf{1.93} \\
    4.8s & 1.49 & 3.10 & \textbf{1.30} & \textbf{2.72} \\ 
    \hline
  \end{tabular}
  }
\end{table}

\subsubsection{Robustness Against Error Drifting over Extended Horizons}
Continuous generative forecasting models often become less reliable as the
prediction horizon grows, partly because small rollout errors can accumulate
over time. To assess this effect, Table~\ref{tab:eth_avg} reports performance
on ETH over extended horizons from $1.2\text{s}$ to $4.8\text{s}$. At
$t=4.8\text{s}$, vanilla MoFlow shows a noticeable increase in
$\mathrm{FDE}_{\mathrm{avg}}$, reaching 3.10, whereas TIGFlow-GRPO achieves a
lower value of 2.72. This gap suggests that the proposed post-training
strategy helps maintain more stable long-horizon predictions.

\begin{table}
    \centering
    \caption{Collision rates (\%) across datasets and horizons. TIGFlow-GRPO demonstrates superior collision avoidance in most settings.}
    \label{tab:col_rate}
    \vspace{-0.5em}
    \resizebox{\linewidth}{!}{
    \begin{tabular}{cc|cccccc}
    \midrule
    \textbf{Method} & \textbf{Horizon} & \textbf{ETH} & \textbf{HOTEL} & \textbf{UNIV} & \textbf{ZARA1} & \textbf{ZARA2} & \textbf{SDD} \\
    \midrule
    \multirow{4}{*}{MoFlow~\cite{fu2025moflow}} & 1.2s & \textbf{0.55} & 2.83 & 5.72 & 1.14 & 2.63 & 1.92 \\
               & 2.4s & 3.55 & 4.74 & 13.23 & 3.64 & 8.12 & 12.62 \\
               & 3.6s & 4.92 & 7.81 & 15.92 & 7.94 & 13.26 & 15.85 \\
               & 4.8s & 5.63 & 9.63 & 19.65 & 11.63 & 16.54 & 19.74 \\
    \midrule
    \multirow{4}{*}{TIGFlow-GRPO} & 1.2s & 0.61 & \textbf{1.35} & \textbf{4.34} & \textbf{0.43} & \textbf{1.58} & \textbf{0.83} \\
                 & 2.4s & \textbf{2.95} & \textbf{2.67} & \textbf{11.04} & \textbf{2.93} & \textbf{6.57} & \textbf{9.45} \\
                 & 3.6s & \textbf{4.53} & \textbf{4.23} & \textbf{14.65} & \textbf{5.84} & \textbf{10.52} & \textbf{11.68} \\
                 & 4.8s & \textbf{5.16} & \textbf{4.95} & \textbf{16.58} & \textbf{7.55} & \textbf{12.03} & \textbf{12.29} \\
    \bottomrule
    \end{tabular}
    }
    \vspace{-0.8em}
\end{table}

\subsubsection{Collision Avoidance Performance}
Table~\ref{tab:col_rate} reports $\mathrm{Col}$ across datasets and forecasting
horizons. Compared with MoFlow, TIGFlow-GRPO achieves lower collision rates in
23 of the 24 dataset-horizon settings. The gains become more pronounced at
longer horizons, where accumulated trajectory errors are more likely to lead to
unsafe interactions. At the $4.8\text{s}$ horizon, $\mathrm{Col}$ decreases
from 19.74\% to 12.29\% on SDD, from 9.63\% to 4.95\% on HOTEL, and from
16.54\% to 12.03\% on ZARA2. Averaged over all dataset-horizon pairs,
$\mathrm{Col}$ drops from 8.72\% to 6.45\%, indicating improved social
compliance beyond what displacement errors alone can reflect. The only minor
exception occurs on ETH at the $1.2\text{s}$ horizon, where the two results are
nearly identical at 0.61\% and 0.55\%. Overall, these results show that
Flow-GRPO consistently suppresses collision-prone samples and yields safer
trajectory distributions, especially for longer-horizon forecasting.

\begin{figure*}[t!]
\centering
\includegraphics[width=\textwidth]{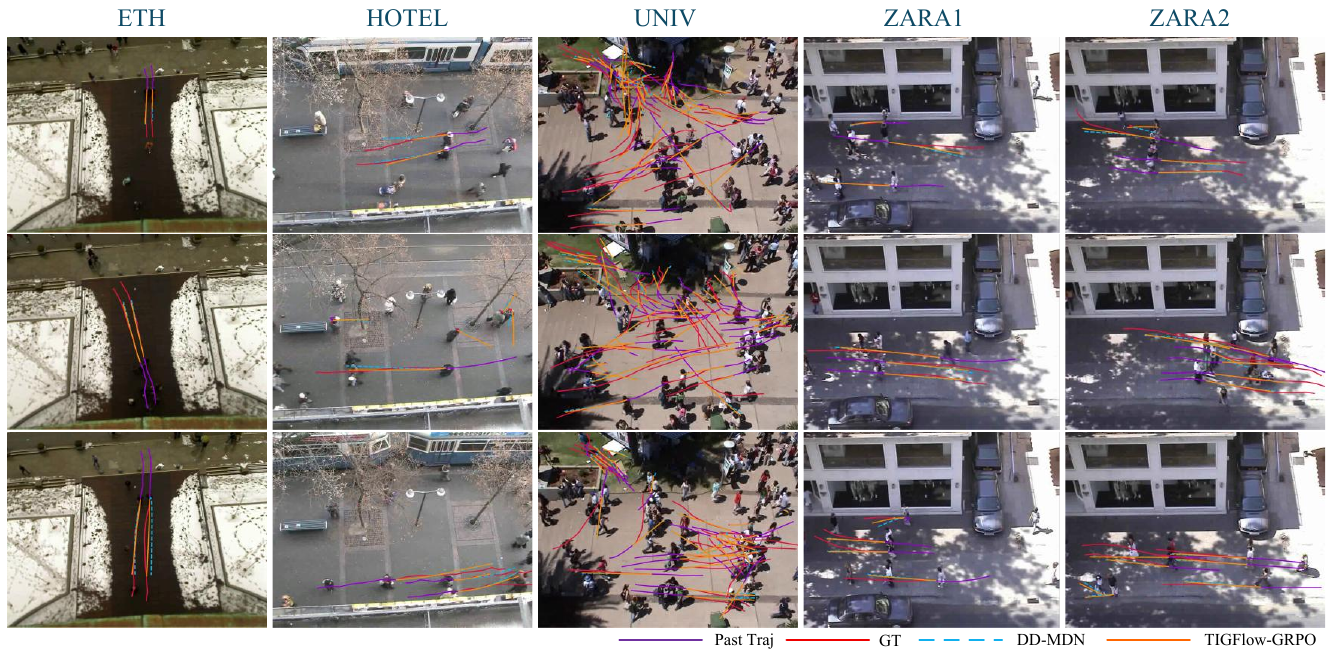}
\caption{Qualitative results on the ETH/UCY benchmark. Compared with DD-MDN,
TIGFlow-GRPO produces predictions that better follow local interaction
patterns and feasible motion paths across ETH, HOTEL, UNIV, ZARA1, and
ZARA2.}
\Description{A 3-by-5 grid of bird's-eye-view trajectory prediction examples
from the ETH/UCY benchmark, with columns corresponding to ETH, HOTEL, UNIV,
ZARA1, and ZARA2. Purple lines denote observed past trajectories, red lines
denote ground-truth futures, yellow lines denote DD-MDN predictions, and
orange lines denote TIGFlow-GRPO predictions. In crowded scenes such as UNIV
and ZARA1, TIGFlow-GRPO produces more interaction-consistent futures. In
structured scenes such as ETH and HOTEL, its predictions better follow
feasible walking regions and scene boundaries.}
\label{fig:visualization}
\end{figure*}
  
\subsection{Ablation Study}
To assess the contribution of each component, we conduct an ablation study on
ETH. We report both minimum and average metrics to better reflect best-case
accuracy and overall distribution quality. As shown in
Table~\ref{tab:ablation}, removing any core component causes a clear
performance drop, with larger degradations in the average metrics, which are
more sensitive to low-quality or outlier trajectories.

\subsubsection{Effectiveness of TIG-GAT}
When the Trajectory-Interaction-Graph Attention (TIG-GAT) module is removed,
the model loses explicit local interaction modeling. As a result, although the
increase in $\mathrm{FDE}_{\min}$ is relatively modest,
$\mathrm{FDE}_{\mathrm{avg}}$ rises more noticeably from 2.718 to 3.103. This
pattern suggests that without explicit interaction encoding, the model is more
likely to generate trajectories that deviate from the local motion context,
thereby increasing low-quality outliers in the sample distribution.

\begin{table}
    \centering
    \caption{Ablation study exemplified on the ETH dataset. \textbf{Bold} values indicate the best performance. The removal of any core component noticeably degrades performance.}
    \label{tab:ablation}
    \renewcommand{\arraystretch}{1.2} 
    \setlength{\tabcolsep}{4pt} 
    \setlength{\aboverulesep}{0pt}
    \setlength{\belowrulesep}{0pt}
    \vspace{-0.5em}
    \resizebox{\linewidth}{!}{
    \begin{tabular}{c|cccc}
    \toprule
    \raisebox{-2.0ex}{Model Variant} & \multicolumn{4}{c}{\raisebox{-0.5ex}{\textbf{ETH}}} \\
    \cmidrule(lr){2-5} 
    & ADE$_{\text{min}}$ & FDE$_{\text{min}}$ & ADE$_{\text{avg}}$ & FDE$_{\text{avg}}$ \\ 
    \midrule
    w/o TIG-GAT       & 0.404 & 0.593 & 1.494 & 3.103 \\
    w/o social reward & 0.401 & 0.581 & 1.316 & 2.745 \\
    w/o map reward    & 0.398 & 0.577 & 1.301 & 2.719 \\
    w/o GRPO          & 0.394 & 0.555 & 1.328 & 2.855 \\ 
    \midrule
    \rowcolor{gray!20} \textbf{Ours (Full)} & \textbf{0.391} & \textbf{0.553} & \textbf{1.300} & \textbf{2.718} \\ 
    \bottomrule
    \end{tabular}
    }
    \vspace{-1.0em}
\end{table}

\subsubsection{Impact of Social and Map Rewards}
We further ablate the view-aware social reward and the map-aware semantic
reward individually. Removing either reward increases both the minimum and
average displacement errors. This result suggests that supervised pretraining
alone may generate reasonable samples in some cases, yet still fails to
consistently respect interaction patterns or scene constraints across the full
sample set. Explicit social and spatial constraints are therefore important
for shaping the overall trajectory distribution.

\subsubsection{Necessity of GRPO Post-training}
When the entire GRPO post-training stage is removed and inference reverts to
deterministic ODE rollout, $\mathrm{FDE}_{\mathrm{avg}}$ increases to 2.855.
This degradation suggests that supervised flow-matching pretraining alone is
insufficient to control low-quality samples in the predicted distribution. By
contrast, GRPO post-training introduces reward-guided regularization during
rollout, which suppresses outlier trajectories and improves the expected
quality of the sample set. These results suggest that the contribution of GRPO
extends beyond refining individual modes and helps produce a more stable
overall trajectory distribution.

\subsection{Visualization}
Figure~\ref{fig:visualization} presents qualitative comparisons on
representative cases from the five ETH/UCY scenes. In each example, purple
lines denote the observed trajectories, red lines denote the ground-truth
futures, yellow lines denote the predictions of DD-MDN, and orange lines denote
the predictions of TIGFlow-GRPO. Compared with DD-MDN, the proposed method
produces trajectory bundles that are generally tighter and better aligned with
local motion structure. This difference is especially clear in crowded scenes
such as UNIV, where TIGFlow-GRPO maintains multimodal diversity while avoiding
the overly scattered predictions shown by DD-MDN under dense crossing flows. In
ETH and HOTEL, its predictions remain closer to the dominant walking corridor
and exhibit fewer unreasonable lateral deviations. Similar behavior appears in
ZARA1 and ZARA2, where TIGFlow-GRPO better follows sidewalk-like motion
patterns and captures branching futures with more coherent interaction-aware
samples. Overall, the visualizations show that the proposed method generates
multimodal futures that are not only diverse, but also better aligned with
local interactions and scene constraints.

\section{Conclusion}
This paper presented TIGFlow-GRPO, a two-stage approach for human trajectory
forecasting in visually complex scenes. The framework combines a conditional
flow-matching backbone with the TIG-GAT module to capture view-aware social
interactions, and introduces an ODE-to-SDE reformulation for Flow-GRPO
post-training in continuous trajectory space. A composite reward further
integrates social and physical constraints into the alignment process.
Experiments on the ETH/UCY and SDD datasets show that the proposed method
improves multimodal forecasting accuracy and long-horizon robustness, while
generating trajectories that better respect interaction patterns and scene
constraints. Ablation studies further confirm that both TIG-GAT and
reward-guided post-training contribute to these improvements. In future work,
we plan to incorporate richer visual cues and VLM-based spatial-temporal scene reasoning to
support more general motion prediction in intelligent multimedia systems.

\bibliographystyle{ACM-Reference-Format}
\bibliography{main}

@String{Computer = "{IEEE} Computer" }

@String{Chelsea = "Chelsea" }

@String{Springer = "Springer-Verlag" }

@article{ho2020denoising,
  title={Denoising diffusion probabilistic models},
  author={Ho, Jonathan and Jain, Ajay and Abbeel, Pieter},
  journal={Advances in neural information processing systems},
  volume={33},
  pages={6840--6851},
  year={2020}
}

@article{dhariwal2021diffusion,
  title={Diffusion models beat gans on image synthesis},
  author={Dhariwal, Prafulla and Nichol, Alexander},
  journal={Advances in neural information processing systems},
  volume={34},
  pages={8780--8794},
  year={2021}
}

@article{lipman2022flow,
  title={Flow matching for generative modeling},
  author={Lipman, Yaron and Chen, Ricky TQ and Ben-Hamu, Heli and Nickel, Maximilian and Le, Matt},
  journal={arXiv preprint arXiv:2210.02747},
  year={2022}
}

@inproceedings{esser2024scaling,
  title={Scaling rectified flow transformers for high-resolution image synthesis},
  author={Esser, Patrick and Kulal, Sumith and Blattmann, Andreas and Entezari, Rahim and M{\"u}ller, Jonas and Saini, Harry and Levi, Yam and Lorenz, Dominik and Sauer, Axel and Boesel, Frederic and others},
  booktitle={Forty-first international conference on machine learning},
  year={2024}
}

@article{stiennon2020learning,
  title={Learning to summarize with human feedback},
  author={Stiennon, Nisan and Ouyang, Long and Wu, Jeffrey and Ziegler, Daniel and Lowe, Ryan and Voss, Chelsea and Radford, Alec and Amodei, Dario and Christiano, Paul F},
  journal={Advances in neural information processing systems},
  volume={33},
  pages={3008--3021},
  year={2020}
}

@article{gat2024discrete,
  title={Discrete flow matching},
  author={Gat, Itai and Remez, Tal and Shaul, Neta and Kreuk, Felix and Chen, Ricky TQ and Synnaeve, Gabriel and Adi, Yossi and Lipman, Yaron},
  journal={Advances in Neural Information Processing Systems},
  volume={37},
  pages={133345--133385},
  year={2024}
}

@article{polyak2024movie,
  title={Movie gen: A cast of media foundation models},
  author={Polyak, Adam and Zohar, Amit and Brown, Andrew and Tjandra, Andros and Sinha, Animesh and Lee, Ann and Vyas, Apoorv and Shi, Bowen and Ma, Chih-Yao and Chuang},
  journal={arXiv preprint arXiv:2410.13720},
  year={2024}
}

@article{liu2022flow,
  title={Flow straight and fast: Learning to generate and transfer data with rectified flow},
  author={Liu, Xingchao and Gong, Chengyue and Liu, Qiang},
  journal={arXiv preprint arXiv:2209.03003},
  year={2022}
}

@article{liu2025flow,
  title={Flow-grpo: Training flow matching models via online rl},
  author={Liu, Jie and Liu, Gongye and Liang, Jiajun and Li, Yangguang and Liu, Jiaheng and Wang, Xintao and Wan, Pengfei and Zhang, Di and Ouyang, Wanli},
  journal={arXiv preprint arXiv:2505.05470},
  year={2025}
}

@article{ouyang2022training,
  title={Training language models to follow instructions with human feedback},
  author={Ouyang, Long and Wu, Jeffrey and Jiang, Xu and Almeida, Diogo and Wainwright, Carroll and Mishkin, Pamela and Zhang, Chong and Agarwal, Sandhini and Slama, Katarina and Ray, Alex and others},
  journal={Advances in neural information processing systems},
  volume={35},
  pages={27730--27744},
  year={2022}
}

@inproceedings{fan2025online,
  title={Online reward-weighted fine-tuning of flow matching with wasserstein regularization},
  author={Fan, Jiajun and Shen, Shuaike and Cheng, Chaoran and Chen, Yuxin and Liang, Chumeng and Liu, Ge},
  booktitle={The Thirteenth International Conference on Learning Representations},
  year={2025}
}

@article{zheng2025yingmusic,
  title={YingMusic-Singer: Zero-shot Singing Voice Synthesis and Editing with Annotation-free Melody Guidance},
  author={Zheng, Junjie and Hao, Chunbo and Ma, Guobin and Zhang, Xiaoyu and Chen, Gongyu and Ding, Chaofan and Chen, Zihao and Xie, Lei},
  journal={arXiv preprint arXiv:2512.04779},
  year={2025}
}

@inproceedings{fan2023reinforcement,
  title={Reinforcement learning for fine-tuning text-to-image diffusion models},
  author={Fan, Ying and Watkins, Olivia and Du, Yuqing and Liu, Hao and Ryu, Moonkyung and Boutilier, Craig and Abbeel, Pieter and Ghavamzadeh, Mohammad and Lee, Kangwook and Lee, Kimin},
  booktitle={Thirty-seventh Conference on Neural Information Processing Systems (NeurIPS) 2023},
  year={2023},
  organization={Neural Information Processing Systems Foundation}
}

@article{shao2024deepseekmath,
  title={Deepseekmath: Pushing the limits of mathematical reasoning in open language models},
  author={Shao, Zhihong and Wang, Peiyi and Zhu, Qihao and Xu, Runxin and Song, Junxiao and Bi, Xiao and Zhang, Haowei and Zhang, Mingchuan and Li, YK and Wu, Yang and others},
  journal={arXiv preprint arXiv:2402.03300},
  year={2024}
}

@article{song2020score,
  title={Score-based generative modeling through stochastic differential equations},
  author={Song, Yang and Sohl-Dickstein, Jascha and Kingma, Diederik P and Kumar, Abhishek and Ermon, Stefano and Poole, Ben},
  journal={arXiv preprint arXiv:2011.13456},
  year={2020}
}

@article{albergo2025stochastic,
  title={Stochastic interpolants: A unifying framework for flows and diffusions},
  author={Albergo, Michael and Boffi, Nicholas M and Vanden-Eijnden, Eric},
  journal={Journal of Machine Learning Research},
  volume={26},
  number={209},
  pages={1--80},
  year={2025}
}

@article{domingo2024adjoint,
  title={Adjoint matching: Fine-tuning flow and diffusion generative models with memoryless stochastic optimal control},
  author={Domingo-Enrich, Carles and Drozdzal, Michal and Karrer, Brian and Chen, Ricky TQ},
  journal={arXiv preprint arXiv:2409.08861},
  year={2024}
}

@article{xue2025advantage,
  title={Advantage weighted matching: Aligning rl with pretraining in diffusion models},
  author={Xue, Shuchen and Ge, Chongjian and Zhang, Shilong and Li, Yichen and Ma, Zhi-Ming},
  journal={arXiv preprint arXiv:2509.25050},
  year={2025}
}

@article{sun2025f5r,
  title={F5r-tts: Improving flow-matching based text-to-speech with group relative policy optimization},
  author={Sun, Xiaohui and Xiao, Ruitong and Mo, Jianye and Wu, Bowen and Yu, Qun and Wang, Baoxun},
  journal={arXiv preprint arXiv:2504.02407},
  year={2025}
}

@inproceedings{dong2025leveraging,
  title={Leveraging sd map to augment hd map-based trajectory prediction},
  author={Dong, Zhiwei and Ding, Ran and Li, Wei and Zhang, Peng and Tang, Guobin and Guo, Jia},
  booktitle={Proceedings of the Computer Vision and Pattern Recognition Conference},
  pages={17219--17228},
  year={2025}
}

@inproceedings{xiao2025srefiner,
  title={SRefiner: Soft-Braid Attention for Multi-Agent Trajectory Refinement},
  author={Xiao, Liwen and Pan, Zhiyu and Wang, Zhicheng and Cao, Zhiguo and Li, Wei},
  booktitle={Proceedings of the IEEE/CVF International Conference on Computer Vision},
  pages={960--969},
  year={2025}
}

@inproceedings{bae2023eigentrajectory,
  title={Eigentrajectory: Low-rank descriptors for multi-modal trajectory forecasting},
  author={Bae, Inhwan and Oh, Jean and Jeon, Hae-Gon},
  booktitle={Proceedings of the IEEE/CVF International Conference on Computer Vision},
  pages={10017--10029},
  year={2023}
}

@inproceedings{bae2024singulartrajectory,
  title={Singulartrajectory: Universal trajectory predictor using diffusion model},
  author={Bae, Inhwan and Park, Young-Jae and Jeon, Hae-Gon},
  booktitle={Proceedings of the IEEE/CVF Conference on Computer Vision and Pattern Recognition},
  pages={17890--17901},
  year={2024}
}

@inproceedings{bahari2025certified,
  title={Certified human trajectory prediction},
  author={Bahari, Mohammadhossein and Saadatnejad, Saeed and Farsangi, Amirhossein Askari and Moosavi-Dezfooli, Seyed-Mohsen and Alahi, Alexandre},
  booktitle={Proceedings of the Computer Vision and Pattern Recognition Conference},
  pages={12301--12311},
  year={2025}
}

@inproceedings{fang2025neuralized,
  title={Neuralized Markov Random Field for Interaction-Aware Stochastic Human Trajectory Prediction.},
  author={Fang, Zilin and Hsu, David and Lee, Gim Hee and Lee, Gim Hee},
  booktitle={ICLR},
  year={2025}
}

@inproceedings{fu2025moflow,
  title={Moflow: One-step flow matching for human trajectory forecasting via implicit maximum likelihood estimation based distillation},
  author={Fu, Yuxiang and Yan, Qi and Wang, Lele and Li, Ke and Liao, Renjie},
  booktitle={Proceedings of the Computer Vision and Pattern Recognition Conference},
  pages={17282--17293},
  year={2025}
}

@inproceedings{gao2025socialmp,
  title={SocialMP: Learning Social Aware Motion Patterns via Additive Fusion for Pedestrian Trajectory Prediction},
  author={Gao, Tianci and Zhang, Yuzhen and Guo, Hang and Lv, Pei},
  booktitle={Proceedings of the Thirty-Fourth International Joint Conference on Artificial Intelligence},
  pages={90--98},
  year={2025}
}

@inproceedings{gu2022stochastic,
  title={Stochastic trajectory prediction via motion indeterminacy diffusion},
  author={Gu, Tianpei and Chen, Guangyi and Li, Junlong and Lin, Chunze and Rao, Yongming and Zhou, Jie and Lu, Jiwen},
  booktitle={Proceedings of the IEEE/CVF conference on computer vision and pattern recognition},
  pages={17113--17122},
  year={2022}
}

@article{hetzel2026dd,
  title={DD-MDN: Human Trajectory Forecasting with Diffusion-Based Dual Mixture Density Networks and Uncertainty Self-Calibration},
  author={Hetzel, Manuel and Turacan, Kerim and Reichert, Hannes and Doll, Konrad and Sick, Bernhard},
  journal={arXiv preprint arXiv:2602.11214},
  year={2026}
}

@inproceedings{jeong2025multi,
  title={Multi-modal knowledge distillation-based human trajectory forecasting},
  author={Jeong, Jaewoo and Lee, Seohee and Park, Daehee and Lee, Giwon and Yoon, Kuk-Jin},
  booktitle={Proceedings of the Computer Vision and Pattern Recognition Conference},
  pages={24222--24233},
  year={2025}
}

@article{liu2025egotraj,
  title={EgoTraj-Bench: Towards Robust Trajectory Prediction Under Ego-view Noisy Observations},
  author={Liu, Jiayi and Zhou, Jiaming and Ye, Ke and Lin, Kun-Yu and Wang, Allan and Liang, Junwei},
  journal={arXiv preprint arXiv:2510.00405},
  year={2025}
}

@inproceedings{salzmann2020trajectron++,
  title={Trajectron++: Dynamically-feasible trajectory forecasting with heterogeneous data},
  author={Salzmann, Tim and Ivanovic, Boris and Chakravarty, Punarjay and Pavone, Marco},
  booktitle={European conference on computer vision},
  pages={683--700},
  year={2020},
  organization={Springer}
}

@inproceedings{xu2022adaptive,
  title={Adaptive trajectory prediction via transferable gnn},
  author={Xu, Yi and Wang, Lichen and Wang, Yizhou and Fu, Yun},
  booktitle={Proceedings of the IEEE/CVF conference on computer vision and pattern recognition},
  pages={6520--6531},
  year={2022}
}

@inproceedings{xu2022groupnet,
  title={Groupnet: Multiscale hypergraph neural networks for trajectory prediction with relational reasoning},
  author={Xu, Chenxin and Li, Maosen and Ni, Zhenyang and Zhang, Ya and Chen, Siheng},
  booktitle={Proceedings of the IEEE/CVF conference on computer vision and pattern recognition},
  pages={6498--6507},
  year={2022}
}

@inproceedings{xu2022socialvae,
  title={Socialvae: Human trajectory prediction using timewise latents},
  author={Xu, Pei and Hayet, Jean-Bernard and Karamouzas, Ioannis},
  booktitle={European Conference on Computer Vision},
  pages={511--528},
  year={2022},
  organization={Springer}
}

@inproceedings{xu2023eqmotion,
  title={Eqmotion: Equivariant multi-agent motion prediction with invariant interaction reasoning},
  author={Xu, Chenxin and Tan, Robby T and Tan, Yuhong and Chen, Siheng and Wang, Yu Guang and Wang, Xinchao and Wang, Yanfeng},
  booktitle={Proceedings of the IEEE/CVF conference on computer vision and pattern recognition},
  pages={1410--1420},
  year={2023}
}

@inproceedings{mao2023leapfrog,
  title={Leapfrog diffusion model for stochastic trajectory prediction},
  author={Mao, Weibo and Xu, Chenxin and Zhu, Qi and Chen, Siheng and Wang, Yanfeng},
  booktitle={Proceedings of the IEEE/CVF conference on computer vision and pattern recognition},
  pages={5517--5526},
  year={2023}
}

@inproceedings{mangalam2021goals,
  title={From goals, waypoints \& paths to long term human trajectory forecasting},
  author={Mangalam, Karttikeya and An, Yang and Girase, Harshayu and Malik, Jitendra},
  booktitle={Proceedings of the IEEE/CVF international conference on computer vision},
  pages={15233--15242},
  year={2021}
}

@inproceedings{gupta2018social,
  title={Social gan: Socially acceptable trajectories with generative adversarial networks},
  author={Gupta, Agrim and Johnson, Justin and Fei-Fei, Li and Savarese, Silvio and Alahi, Alexandre},
  booktitle={Proceedings of the IEEE conference on computer vision and pattern recognition},
  pages={2255--2264},
  year={2018}
}

@inproceedings{hetzel2024reliable,
  title={Reliable probabilistic human trajectory prediction for autonomous applications},
  author={Hetzel, Manuel and Reichert, Hannes and Doll, Konrad and Sick, Bernhard},
  booktitle={European Conference on Computer Vision},
  pages={135--152},
  year={2024},
  organization={Springer}
}

@inproceedings{jiang2023motiondiffuser,
  title={Motiondiffuser: Controllable multi-agent motion prediction using diffusion},
  author={Jiang, Chiyu and Cornman, Andre and Park, Cheolho and Sapp, Benjamin and Zhou, Yin and Anguelov, Dragomir and others},
  booktitle={Proceedings of the IEEE/CVF conference on computer vision and pattern recognition},
  pages={9644--9653},
  year={2023}
}

@article{vaswani2017attention,
  title={Attention is all you need},
  author={Vaswani, Ashish and Shazeer, Noam and Parmar, Niki and Uszkoreit, Jakob and Jones, Llion and Gomez, Aidan N and Kaiser, {\L}ukasz and Polosukhin, Illia},
  journal={Advances in neural information processing systems},
  volume={30},
  year={2017}
}

@inproceedings{pellegrini2009you,
  title={You'll never walk alone: Modeling social behavior for multi-target tracking},
  author={Pellegrini, Stefano and Ess, Andreas and Schindler, Konrad and Van Gool, Luc},
  booktitle={2009 IEEE 12th international conference on computer vision},
  pages={261--268},
  year={2009},
  organization={IEEE}
}

@inproceedings{lerner2007crowds,
  title={Crowds by example},
  author={Lerner, Alon and Chrysanthou, Yiorgos and Lischinski, Dani},
  booktitle={Computer graphics forum},
  volume={26},
  number={3},
  pages={655--664},
  year={2007},
  organization={Wiley Online Library}
}

@inproceedings{robicquet2016learning,
  title={Learning social etiquette: Human trajectory understanding in crowded scenes},
  author={Robicquet, Alexandre and Sadeghian, Amir and Alahi, Alexandre and Savarese, Silvio},
  booktitle={European conference on computer vision},
  pages={549--565},
  year={2016},
  organization={Springer}
}

\end{document}